\documentclass[10pt,twocolumn,letterpaper]{article}
\usepackage{cvpr}
\usepackage{times}
\usepackage{epsfig}
\usepackage{graphicx}
\usepackage{amsmath}
\usepackage{amssymb}
\usepackage{cuted}
\usepackage{caption}
\usepackage{mathtools}
\usepackage{wrapfig}
\usepackage{xcolor}
\DeclarePairedDelimiterX{\norm}[1]{\lVert}{\rVert}{#1}
\DeclareMathOperator*{\argmin}{\arg\!\min}

\usepackage{environ}
\NewEnviron{myequation}{%
\begin{equation}
\scalebox{3}{$\BODY$}
\end{equation}
}

\usepackage{afterpage}
\usepackage{everypage}
\makeatletter
\newif\ifdelaymatch

\newcommand{\delayfloat}[4]{
  \ifnum\value{page}<#2\relax
    \afterpage{\delayfloat{#1}{#2}{#3}{#4}}%
  \else
    \delaymatchfalse
    \ifcase#3\relax\or
      \if@firstcolumn \delaymatchtrue \fi
    \or
      \if@firstcolumn\else \delaymatchtrue \fi
    \fi
    \ifdelaymatch
      \begin{#1}[t]
        \box#4
      \end{#1}
    \else
      \afterpage{\delayfloat{#1}{#2}{#3}{#4}}%
    \fi
  \fi}

\newenvironment{delayed}[4]{
  \ifcase#3\relax
    \errmessage{Bad delayed column argument: #3 (must be 1 or 2)}%
  \or\or\else
    \errmessage{Bad delayed column argument: #3 (must be 1 or 2)}%
  \fi
  \def\delayed@box{#4}
  \def\delayed@args{{#1}{#2}{#3}{#4}}%
  \begin{lrbox}{#4}\begin{minipage}{\columnwidth}%
    \def\@captype{#1}%
}{
  \end{minipage}\end{lrbox}%
  \global\setbox\delayed@box=\copy\delayed@box
  \expandafter\delayfloat\delayed@args
}

\newif\ifdelaymatchtop

\newcommand{\delaytop}[2]{
  \ifnum\value{page}<#1\relax
    \afterpage{\delaytop{#1}{#2}}%
  \else
    \twocolumn[\box#2\par\vskip\dbltextfloatsep]%
  \fi}

\newenvironment{delayedtop*}[3]{
  \def\delayedtop@box{#3}
  \def\delayedtop@args{{#2}{#3}}%
  \begin{lrbox}{#3}\begin{minipage}{\textwidth}%
    \def\@captype{#1}%
}{
  \end{minipage}\end{lrbox}%
  \global\setbox\delayedtop@box=\copy\delayedtop@box
  \expandafter\delaytop\delayedtop@args
}

\makeatother

\newsavebox{\tempboxA}
\newsavebox{\tempboxARC}
\newsavebox{\tempboxARCC}
\newsavebox{\tempboxB}
\newsavebox{\tempboxC}
\newsavebox{\tempboxCC}
\newsavebox{\tempboxD}
\newsavebox{\tempboxE}
\newsavebox{\tempboxF}
\newsavebox{\tempboxG}
\newsavebox{\tempboxTOSCA}
\newsavebox{\tempboxSCAPEgeo}
\newsavebox{\tempboxRealScans}
\newsavebox{\tempboxPartial}

\usepackage[draft=false,pagebackref=true,breaklinks=true,letterpaper=true,colorlinks,bookmarks=false]{hyperref}

\cvprfinalcopy 

\def\cvprPaperID{3145} 

\ifcvprfinal\fi
\begin{document}
\begin{delayed}{figure}{1}{2}{\tempboxA}
\centering
  
  \includegraphics[width=\linewidth]{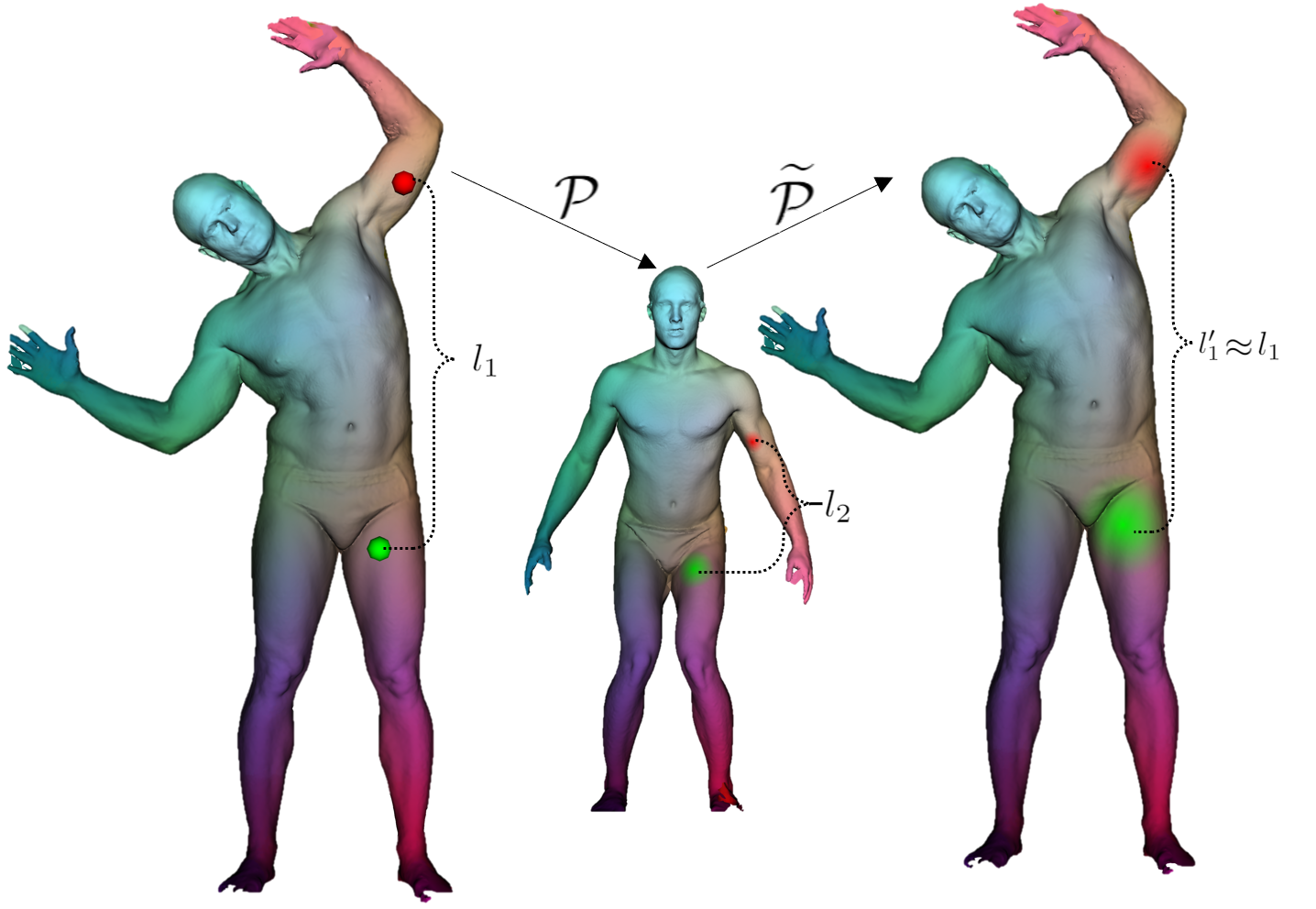}
  
  \caption{Self-supervised dense correspondence using a cycle mapping architecture. By minimizing the geodesic distortion only on the source shape, we can learn complex
  deformations between structures.}
  \vspace*{-3mm}
  \label{fig:back_forth}
  
\end{delayed}

\begin{delayed}{figure}{2}{2}{\tempboxB}
  \centering
  \includegraphics[width=\linewidth]{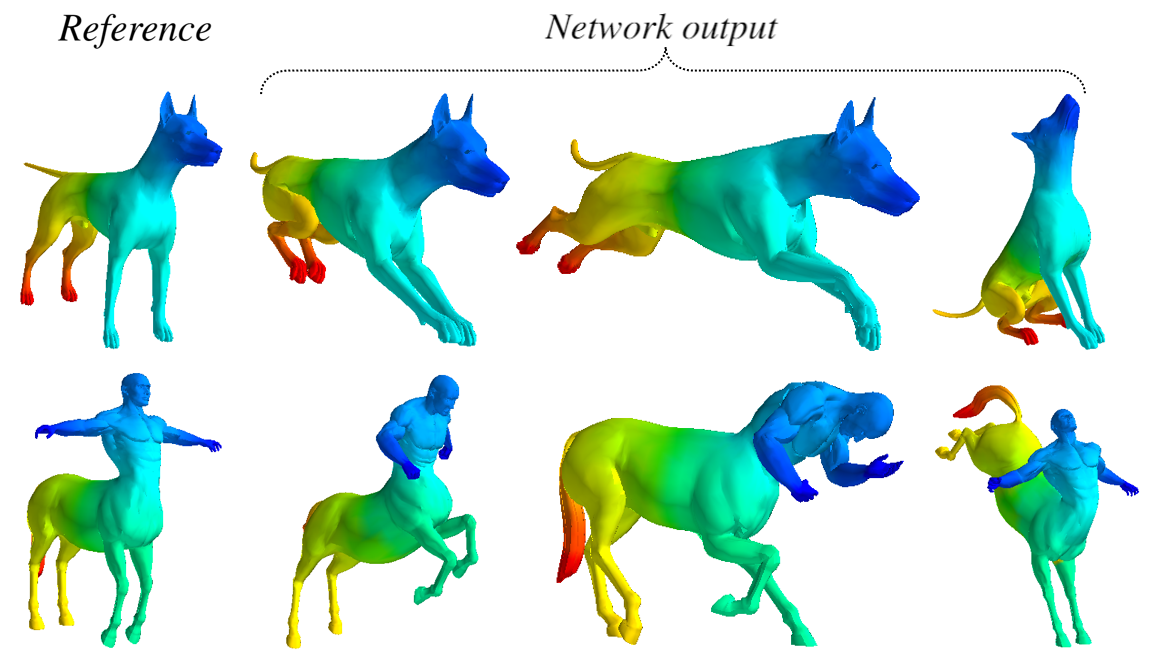}
  \caption{TOSCA dataset results - similar colors represents correspondence mapping - we show excellent  generalization after training for a single epoch on the TOSCA dataset with a pre-trained model on FAUST ~\ref{subsection:tosca}}
  \vspace*{-3mm}
  \label{fig:tosca_visualization}
\end{delayed}

\begin{delayedtop*}{figure}{3}{\tempboxARC}
  \centering
  
  \includegraphics[width=\linewidth]{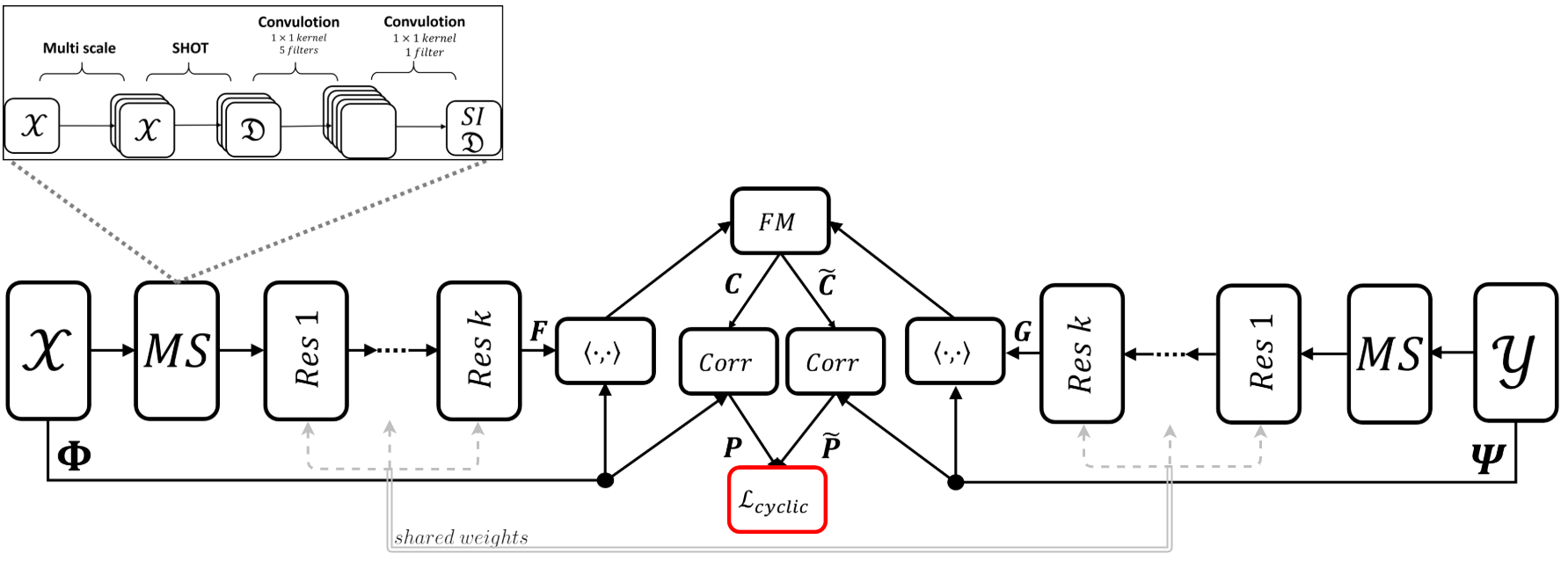}
  \caption{Cyclic functional mapper in between two manifolds $\mathcal{X}$ and $\mathcal{Y}$ (left and right sides). The multi-scaled descriptors (top left, marked $MS$) based on shot~\cite{shotdesc} are passed to a ResNet like network, resulting in two corresponding coefficient matrices $F$ and $G$. By projecting the refined descriptors onto the spectral space, two mappings, $C$ and $\tilde{C}$, are computed. The two soft correspondence matrices $P$ and $\tilde{P}$ are further used as part of the network cyclic loss $\mathcal{L}_{cyclic}$ as shown in equation \ref{eq:cyclic_loss}.}
  \label{fig:architecture_scheme}
\end{delayedtop*}

\begin{delayedtop*}{figure}{4}{\tempboxC}
  \centering
  \includegraphics[width=\linewidth]{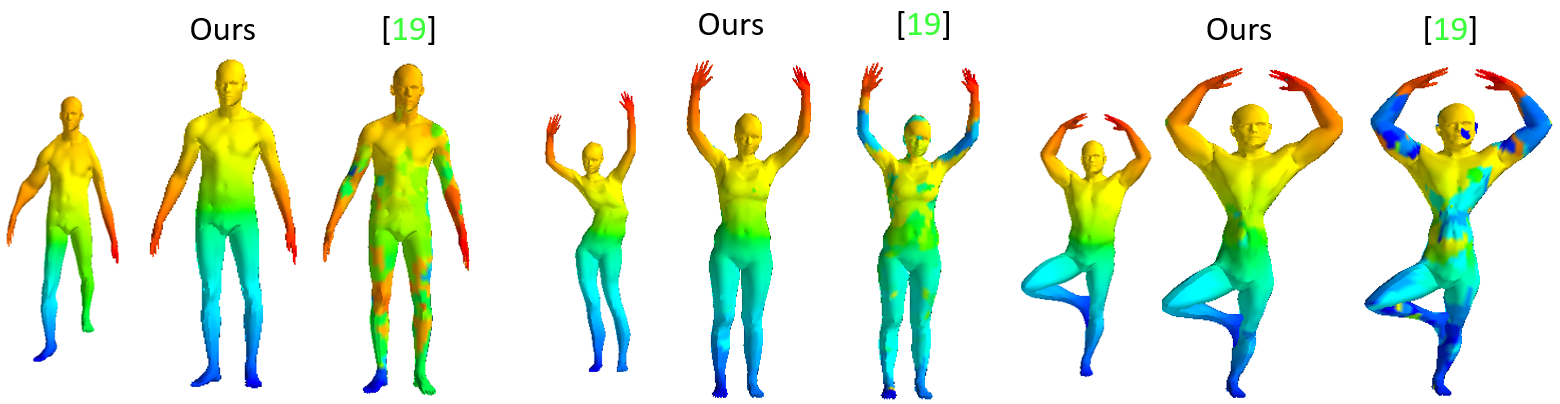}
  \caption{Alignment between non-isometric shapes, where similar parts appear in similar colors. The shapes were locally scaled and stretched while changing their pose. Our approach learns the correct matching while \cite{halimi2019unsupervised} fails under local stretching.} 
  
  \label{fig:local_scaling}
\end{delayedtop*}

\begin{delayed}{table}{5}{2}{\tempboxG}
  \begin{center}
  \begin{tabular}{|l|c|c|c|c|}
    \hline
    Method &
      \multicolumn{2}{c|}{Scans} &
      \multicolumn{2}{c|}{Synthetic} \\
    & inter & intra & inter & intra \\
    \hline
Ours &  \textbf{4.068} & \textbf{2.12} &\textbf{2.327} & \textbf{2.112} \\ 
\hline
    Litany ~\etal \cite{fmnet}& 4.826 & 2.436 &2.452 & 2.125  \\
    \hline
    Halimi ~\etal \cite{halimi2019unsupervised} & 4.883 & 2.51 &3.632 &  2.213 \\
    \hline
    Groueix ~\etal \cite{3d_coded}& 4.833 & 2.789 & --- & --- \\
\hline
    Li ~\etal \cite{lbsauto}& 4.079 & 2.161 & --- & --- \\
\hline
    Chen ~\etal \cite{chen2015robust}& 8.304 & 4.86 & --- & --- \\

    \hline
  \end{tabular}

\end{center}
\vspace*{-5mm}
\caption{Average error on the FAUST dataset measured as distance between mapped points and the ground truth. We compared between our approach and other supervised and unsupervised methods.}
\label{table:real_scal_results}
\end{delayed}

\begin{delayed}{figure}{5}{2}{\tempboxRealScans}
  \centering
  \includegraphics[width=\linewidth]{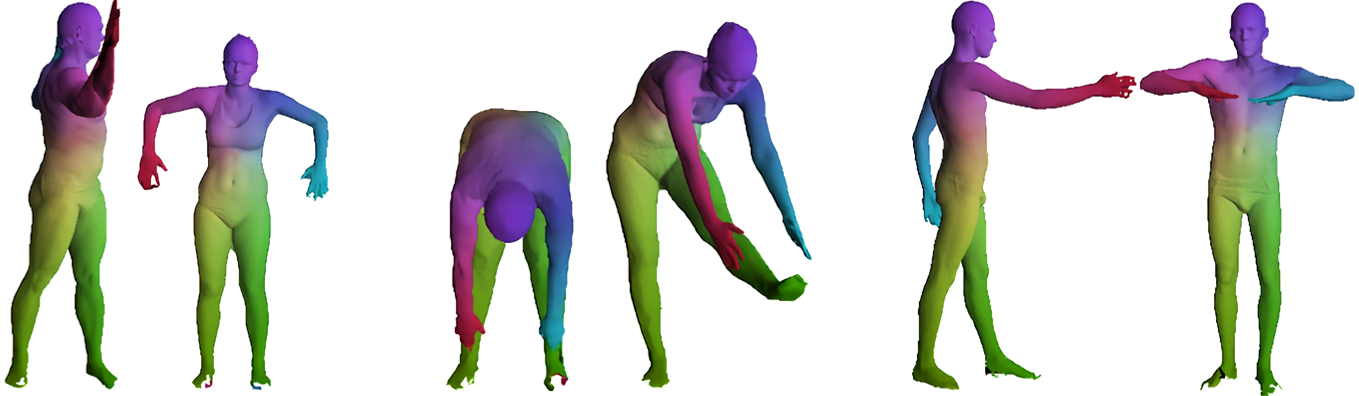}
  \caption{Correspondence on FAUST real scans dataset, where similar colors represent the same correspondence. This dataset contains shapes made of $\sim100k$ vertices with missing information in various poses. We use a post-matching PMF filter \cite{PMF}, and show qualitative results in 
  Table \ref{table:real_scal_results}. We outperform both supervised and unsupervised methods.}
  \vspace*{-3mm}
  \label{fig:faust_real_scans}
\end{delayed}

\begin{delayed}{figure}{6}{2}{\tempboxD}
  \centering
  
  \includegraphics[width=\linewidth]{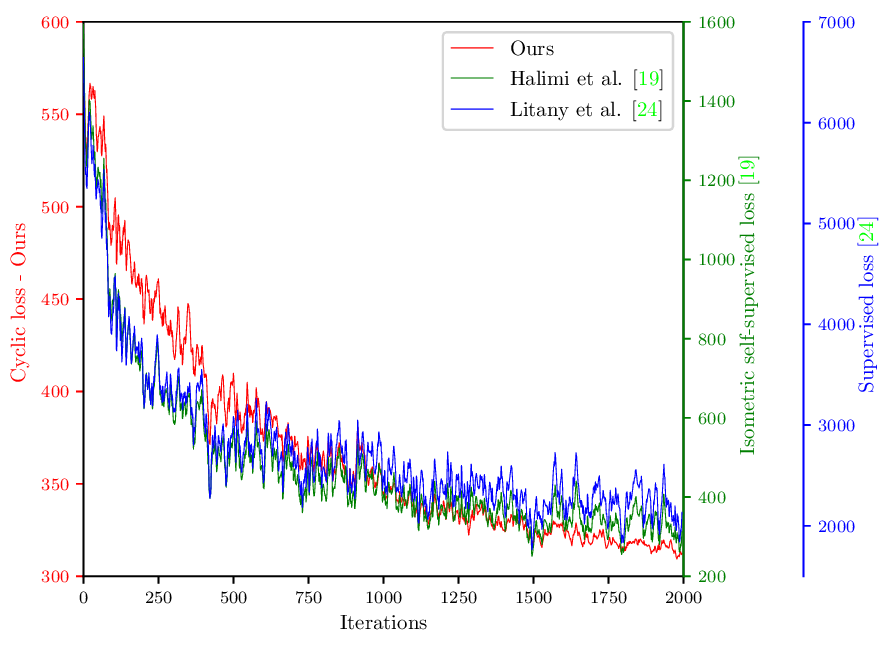}
  \caption{We visualize our cyclic loss, the isometric constrained unsupervised loss \cite{halimi2019unsupervised}, and the supervised loss \cite{fmnet} during the training of our cyclic loss on the synthetic FAUST dataset. We show that minimization of the cyclic loss on isometric structures is equivalent (after normalization) to minimizing the isometric constraint or applying ground truth labels to the learning. }
  \vspace*{-3mm}
  \label{fig:different_losses}
\end{delayed}

\begin{delayed}{figure}{6}{2}{\tempboxE}
  \centering
  \includegraphics[width=\linewidth]{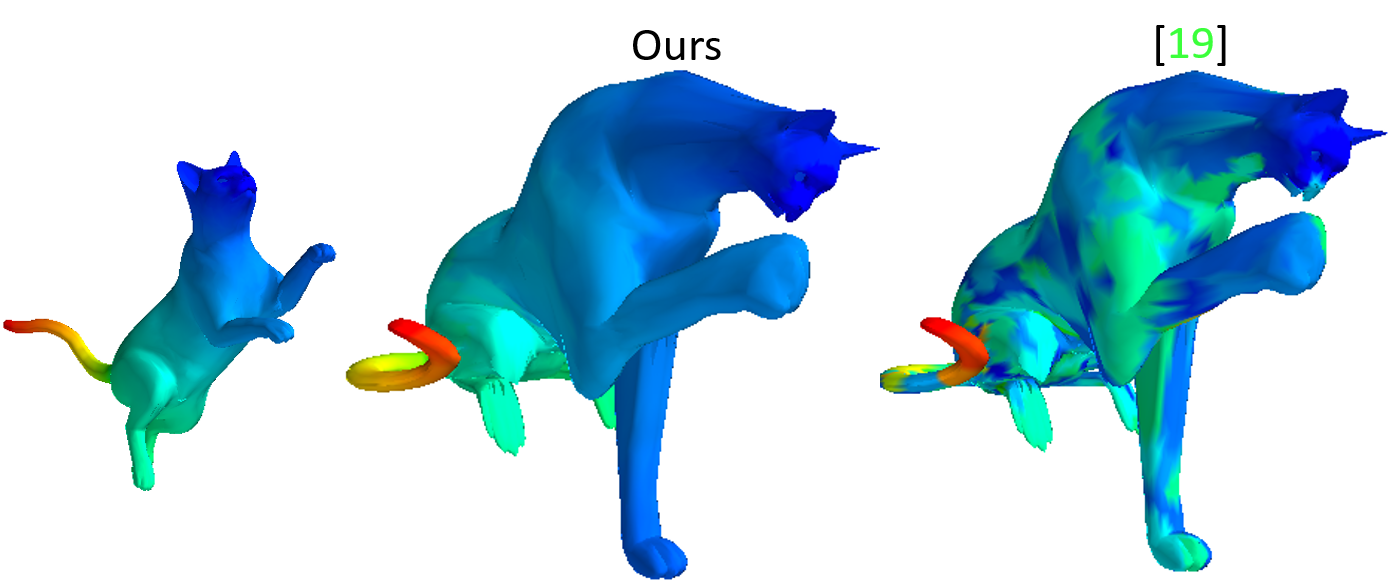}
  \caption{Single pair (one shot) learning on deformable non-isometric shapes. Supervised methods as \cite{fmnet} are irrelevant, where isometric self-learning approach fails \cite{halimi2019unsupervised}.}
  \vspace*{-3mm}
  \label{fig:single_pair_learning}
\end{delayed}

\begin{delayed}{figure}{7}{1}{\tempboxTOSCA}
  \centering
  \includegraphics[width=\linewidth]{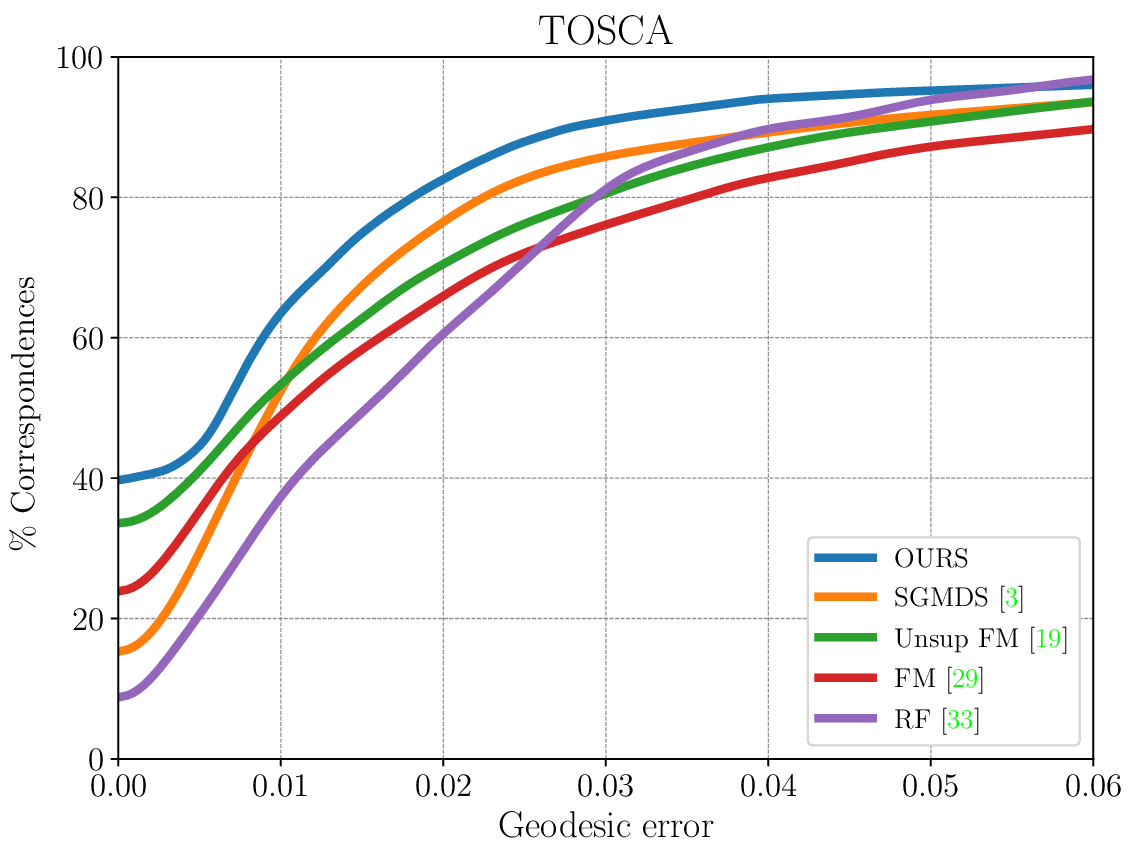}
  \caption{Geodesic error on TOSCA dataset. We report superior results against other supervised and unsupervised learnable methods. Note that the compared methods did not run a post-processing optimization-based filter, or received partial matching as input. }
  \vspace*{-3mm}
  \label{fig:tosca_geo_error}
\end{delayed}

\begin{delayed}{figure}{7}{2}{\tempboxSCAPEgeo}
  \centering
  \includegraphics[width=\linewidth]{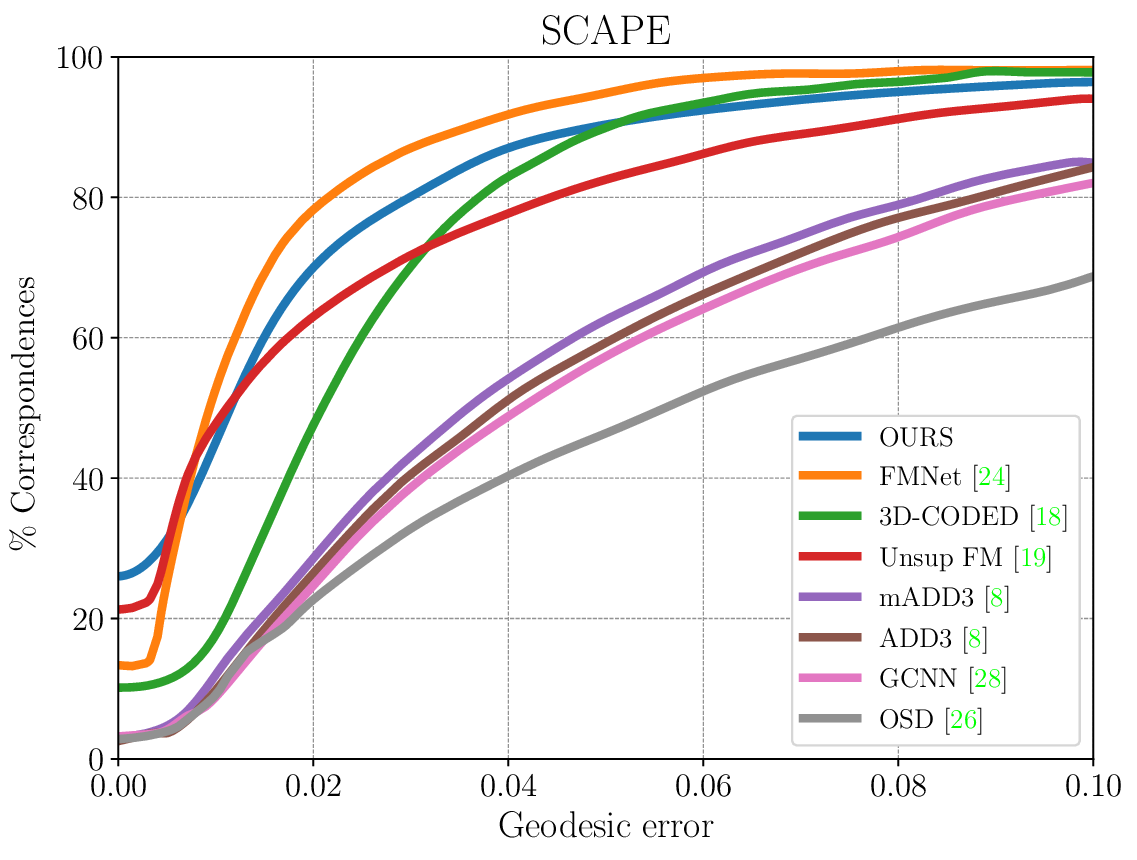}
  \caption{Geodesic error on SCAPE dataset. Our network was trained on FAUST dataset and used to predict the mapping on SCAPE. We provide superior results on all unspervised and almost all supervised methods showing good generalization properties. 
  }
  \vspace*{-3mm}
  \label{fig:scape_geo_error}
\end{delayed}

\begin{delayedtop*}{figure}{8}{\tempboxPartial}
  \centering
  \includegraphics[width=\linewidth]{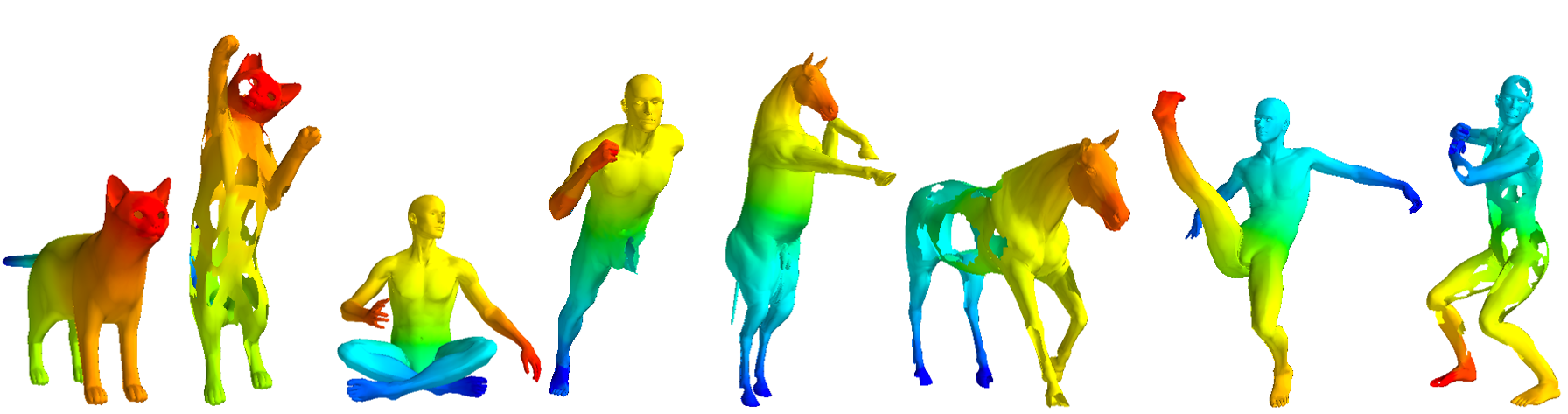}
  \caption{Partial shapes correspondence on SHREC 16 \cite{cosmo2016shrec} dataset after removing substantial parts (upto 75\%).
  In every pair, we mapped the left shape into the right one, where similar mapped points share the color. Our method is robust to missing information.} 
  
  \label{fig:partial_corr}
\end{delayedtop*}


\title{Cyclic Functional Mapping: \protect\\ Self-supervised correspondence  between non-isometric deformable shapes}

\author{Dvir Ginzburg\\
Tel Aviv University\\
{\tt\small dvirginzburg@mail.tau.ac.il}
\and
Dan Raviv\\
Tel Aviv University\\
{\tt\small darav@tauex.tau.ac.il}
}

\maketitle

\begin{abstract}
We present the first utterly self-supervised network for dense correspondence mapping between non-isometric shapes.
The task of alignment in non-Euclidean domains is one of the most fundamental and crucial problems in computer vision. As 3D scanners can generate highly complex and dense models, the mission of finding dense mappings between those models is vital.
The novelty of our solution is based on a cyclic mapping between metric spaces, where the distance between a pair of points should remain invariant after the full cycle. As the same learnable rules that generate the point-wise descriptors apply in both directions, the network learns invariant structures without any labels while coping with non-isometric deformations.
We show here state-of-the-art-results by a large margin for a variety of tasks compared to known self-supervised and supervised methods.
\end{abstract}


\section{Introduction}
Alignment of non-rigid shapes is a fundamental problem in computer vision and plays an important role in multiple applications such as pose transfer, cross-shape texture mapping, 3D body scanning, and simultaneous localization and mapping (SLAM).
The task of finding dense correspondence is especially challenging for non-rigid shapes, as the number of variables needed to define the mapping is vast, and local deformations might occur.
To this end, we have seen a variety of papers focusing on defining unique key-points. These features capture the local uniqueness of the models using curvature ~\cite{curvdesc}, normals ~\cite{shotdesc}, or heat ~\cite{heatdesc}, for example,  and further exploited for finding a dense mapping ~\cite{HKS,WKS}.
\linebreak\indent A different approach used for alignment is based on pair-wise distortions, where angles ~\cite{benhen2008conformal,lipman2011conformal} or distances ~\cite{geodesic_dist_min,geodesic_dist_min_scale} between pairs of points are minimized. Formulating this as a linear ~\cite{linearapprox} or quadratic ~\cite{geodesic_dist_min_scale,QAP1,QAP2} optimization scheme showed a significant enhancement
 but with a painful time complexity even for small models.
 
To confront the challenges in alignment between stretchable shapes we recognize non-metric methods based on conformal mapping \cite{lipman2011conformal}, experimenting with alternative metrics such as scale ~\cite{aflalo2013scale} or affine invariant metrics \cite{raviv2011affine}, or attempts to embed the shapes into a plane or a cone \cite{bronstein2006multigrid,bronstein2006generalized}, for example.
A significant milestone named functional maps \cite{functional_cor} has shown that such a mapping can be performed on the spectral domain, by aligning functions overlaid on top of the shapes.

Recently, a substantial improvement in dense alignment emerged using data-driven solutions, where axiomatic shape models and deformations were replaced by learnable counterparts. Among those methods a highly successfully research direction was based on learning local features overlaid on the vertices of the shapes \cite{fmnet}, where ResNet  \cite{Resnet} like architecture is used to update SHOT descriptors \cite{shotdesc}.

The main challenge new data-driven geometric alignment algorithms need to face is the lack of data to train on or labeled data used for supervised learning. In many cases, the labeled data is expensive to generate or even infeasible to acquire, as seen, for example, in medical imaging.

A recent approach \cite{halimi2019unsupervised} showed that self-supervised learning could be applied for non-rigid alignment between isometric shapes by preserving the pair-wise distance measured on the source and on the target. While showing good results, an isometric limitation is a strong constraint that is irrelevant in many scenarios. On a different note, self-supervised learning was recently addressed in images, where a cyclic mapping between pictures, known as cyclic-GAN,  was introduced \cite{cyclegan,unsupimg2img, unsupdual}. 
The authors showed that given unpaired collection of images from different domains, a cyclic-loss that measures the distortion achieves robust state-of-the-art results in unsupervised learning for domain transfer.

In this work, we claim that one can learn dense correspondence in a self-supervised manner in between non-isometric structures. We present a new learnable cyclic mechanism, where the same model is used both for forward and backward mapping learning to compensate for deformed parts. We measure the pair-wise distance distortion of the cyclic mapping on randomly chosen pair of points only from the source manifold.
We show here state-of-the-art-results by a large margin for a variety of tasks compared to self-supervised and supervised methods, in isometric and non-isometric setups.

\section{Contribution}
We present an unsupervised learning scheme for dense 3D correspondence between shapes, based on learnable self similarities between metric spaces.
The proposed approach has multiple advantages over other known methods. First, there is no need to define a model for the shapes or the deformations; Second, no need for labeled data with dense correspondence mappings. Third, the proposed method can handle isometric or non-isometric deformations and partial matching.
The cyclic mapping approach allows our system to learn the geometric representation of manifolds by feeding it pairs of matching shapes, even without any labels, by measuring a geometric criterion (pair-wise distance) only on the source.

Our main contribution is based on the understanding that a cyclic mapping between metric spaces which follows the same rules, forces the network to learn invariant parts. We built the cyclic mapping using the functional maps framework \cite{functional_cor}, optimizing for a soft correspondence between shapes on the spectral domains by updating a local descriptor per point. The proposed approach can be adapted to any dimension, and here we provide state-of-the-art results on surfaces. We show results that are comparable to supervised learning \cite{fmnet,3d_coded,lbsauto} methods in the rare case we possess dense correspondence labels, and outperforms self-supervised learning approaches \cite{3d_coded,halimi2019unsupervised} when the shapes are isometric. Once the deformations are not isometric, our method stands out, and outperforms other methods by a large margin.

\section{Background}
Our cyclic mapping is built on top of functional maps architecture.  To explain the foundations of this approach, we must elaborate on distance matrices, functional maps and how to weave deep learning into functional maps. Finally, we discuss an isometric unsupervised approach for the alignment task and its limitations, which motivated this work.
\subsection{Riemannian 2-manifolds}
We model 3D shapes as a Riemannian 2-manifold $(\mathcal{X},g)$, where $\mathcal{X}$ is a real smooth manifold, equipped with an inner product $g_p$ on the tangent space $T_p\mathcal{X}$ at each point $p$ that varies smoothly from point to point in the sense that if $\mathcal{U}$ and $\mathcal{V}$ are differentiable vector fields on $\mathcal{X}$, then $p \to g_p(\mathcal{U}|_p, \mathcal{V}|_p)$ is a smooth function.

We equip the manifolds with a distance function $d\mathcal{X}: \mathcal{X}\times \mathcal{X}\to \Re$ induced by the standard volume form $d\mathcal{X}$.
We state the distance matrix $\mathcal{D_X}$, as a square symmetric matrix, represents the manifold's distance function  $d\mathcal{X}$ such that $${\mathcal{D_{X}}}_{ij}=d\mathcal{X}(X_{i},X_{j})$$

\subsection{Functional maps}
Functional maps ~\cite{functional_cor} stands for matching real-valued functions in between manifolds instead of performing a straight forward point matching. Using a spectral basis, one can extract a compact representation for a match on the spectral domain. The clear advantage here is that many natural constraints on the map become linear constraints on the functional map.
Given two manifolds $\mathcal{X}$ and $\mathcal{Y}$, and functional spaces on top  $F(\mathcal{X})$ and $F(\mathcal{Y})$, we can define a functional map using orthogonal bases $\phi$ and $\psi$   

\begin{align}
\begin{split}
\label{eq:functional_cor}
Tf &= T\sum_{i \geq 1}{\langle f,\phi_i \rangle}_{\mathcal{Y}} \phi_i = \sum_{i \geq 1}{\langle f,\phi_i \rangle}_{\mathcal{Y}} T \phi_i \\ & = \sum_{i,j \geq 1}{\langle f,\phi_i \rangle}_{\mathcal{Y}} \underbrace{\langle T \phi_i,\psi_j\rangle_{\mathcal{X}}}_{
          \mathclap{c_{ij}}}\psi_j,
\end{split}
\end{align}
where ~$C ~\in ~\mathbb{R}^{k \times k}$ represents the mapping in between the domains given $k$ matched functions, and every pair of corresponding functions on-top of the manifolds impose a linear constraint on the mapping.
The coefficient matrix $C$ is deeply depended on the choice of the bases $\phi,\psi$, and as shown in prior works ~\cite{functional_cor,partial_maps,halimi2019unsupervised} a good choice for such bases is the Laplacian eigenfunctions of the shapes.

\subsection{Deep functional maps}
Deep functional maps were first introduced in  \cite{fmnet}, where the mapping $C$ in between shapes was refined by learning new local features per point. The authors showed that using a ResNet \cite{Resnet} like architecture on-top of SHOT ~\cite{shotdesc} descriptors, they can revise the local features in such a way that the global mapping is more accurate. The mapping is presented as a soft correspondence matrix  
$P$ where $P_{ji}$ is the probability $\mathcal{X}_i$ corresponds to $\mathcal{Y}_j$.
The loss of the network is based on geodesic distortion between the  corresponding mapping and the ground truth, reading
\begin{align}
\begin{split}
\label{eq:supervised_loss}
\mathcal{L}_{sup}(\mathcal{X,Y}) = \frac{1}{|\mathcal{X}|}\norm[\Bigg]{ \biggl(P\circ(\mathcal{D_Y}\Pi^*)\biggr) }_F^2,
\end{split}
\end{align}
\noindent
where $|\mathcal{X}|$ is the number of vertices of shape $\mathcal{X}$, and if $|\mathcal{X}|=n$, and $|\mathcal{Y}|=m$, then $\Pi^* \in \Re^{m\times n}$ is the ground-truth mapping between $\mathcal{X}$ and $\mathcal{Y}$, $\mathcal{D_Y} \in \Re^{m\times m}$  is the geodesic distance matrix of $\mathcal{Y}$ , $\circ$ denote a point-wise multiplication, and $||_F$ is the Frobenius norm.
For each target vertex, the loss penalizes by the distance between the actual corresponding vertex and the assumed one, multiplied by the amount of certainty the network has in that correspondence. 
Hence the loss is zero if $$ P_{ji} = 1 \Leftrightarrow \Pi^*(\mathcal{X}_i)=\mathcal{Y}_j$$
as $\mathcal{D}(y,y) = 0\  \forall y\in\mathcal{Y}  $.

\subsection{Self-supervised deep functional maps}
The main drawback of deep functional maps is the need for ground truth labels. Obtaining alignment maps for domains such as ours is a strong requirement, and is infeasible in many cases either due to the cost to generate those datasets, or even impractical to collect.
In a recent paper ~\cite{halimi2019unsupervised}, the authors showed that for isometric deformations, we can replace the ground truth requirement with a different geometric criterion based on pair-wise distances. In practice, they married together the Gromov-Hausdorff framework with the deep functional maps architecture. 

The Gromov-Hausdorff distance which measures the distance in between metric spaces, reads
\begin{equation} \label{eq:Gromov_Hausdorff_Distance}
d_{GH}(\mathcal{X,Y})=\frac{1}{2}\inf_{\pi}(dis(\pi)),
\end{equation}
where the infimum is taken over all correspondence distortions of a given mapping $\pi\colon\mathcal{X}\mapsto\mathcal{Y}$.
This distortion can be translated to a pair-wise distance \cite{geodesic_dist_min,geodesic_error_metric} notation,
which was used by \cite{halimi2019unsupervised} as a geometric criterion in the cost function of a deep functional map setup.
Unfortunately, the pair-wise distance constraint is an extreme demand, forcing the models to be isometric, and can not be fulfilled in many practical scenarios.

\section{Cyclic self-supervised deep functional maps}

The main contribution of this paper is the transition from the pair-wise distance comparison between source and target manifolds to a method that only examines the metric in the source manifold. Every pair of distances are mapped to the target and re-mapped back to the source. We use the same model for the forward and backward mapping to avoid a mode collapse, and we measure the distortion once a cyclic map has been completed, forcing the model to learn how to compensate for the deformations.

\subsection{Correspondence distortion}
A mapping $\pi: M \rightarrow N$ between two manifolds generates a pair-wise distortion
\begin{equation}
    dis_\pi(\mathcal{X},\mathcal{Y}) = \sum_{x_1,x_2\in \mathcal{X}} \rho(d_\mathcal{X}(x_1,x_2),d_\mathcal{Y}(\pi(x_1),\pi(x_2)),
\end{equation}
where $\rho$ is usually an $L_p$ norm metric, and $p=2$ is a useful choice of the parameter.

As isometric mapping preserves pair-wise distances,  minimizing the distances between those pairs provides a good metric-depended correspondence. Specifically,

\begin{eqnarray}
\label{eq:isometric-distorion}
 \pi^{iso}(\mathcal{X},\mathcal{Y}) = \argmin_{\pi:\mathcal{X}\rightarrow \mathcal{Y}} dis_\pi(\mathcal{X},\mathcal{Y}).
\end{eqnarray}
\noindent
Solving \ref{eq:isometric-distorion} takes the form of a quadratic assignment problem.
The main drawback of this criterion, as the name suggests, is the isometric assumption. While it is a powerful tool for isometric mappings, natural phenomena do not follow that convention as stretching exists in the data. To overcome those limitations, we present here the cyclic distortion criterion.

\subsection{Cyclic distortion}
We define a {\bf cyclic distortion} $\pi^{cyc}$ as a composition of two mappings $\pi_{\to}:\mathcal{X} \rightarrow \mathcal{Y}$ and $\pi_{\gets}:\mathcal{Y} \rightarrow \mathcal{X}$, which leads to a cyclic distortion
\begin{eqnarray}
    &&dis_{(\pi_{\to},\pi_{\gets})}^{cyc}(\mathcal{X},\mathcal{Y}) = \\
    &&\sum_{x_1,x_2\in \mathcal{X}} \rho(d_\mathcal{X}(x_1,x_2),d_\mathcal{X}(\tilde x_1,\tilde x_2)),\nonumber
\end{eqnarray}
where $\tilde x_1 =  \pi_{\gets}(\pi_{\to}(x_1)$ and $\tilde x_2=\pi_{\gets}(\pi_{\to}(x_2))$.

$\pi_{\to}$ and $\pi_{\gets}$ are being optimized using the same sub-network, implemented as shared weights in the learning process. Every forward mapping $\pi_{\to}$ induce a backward mapping $\pi_{\gets}$ and vise-versa. We call this coupled pair  $\pi=(\pi_{\to},\pi_{\gets})$ a \emph{conjugate mapping}, and denote the space of all conjugate mappings by $\mathcal {S}$.
We define the cyclic mapping as
\begin{eqnarray}
\label{eq:cyclic-mapping}
 \pi^{cyc}(\mathcal{X},\mathcal{Y}) = \argmin_{ \pi:(\mathcal{X}\rightarrow \mathcal{Y}, \mathcal{Y}\rightarrow \mathcal{X}) \in \mathcal{S}} dis_{\pi}^{cyc}(\mathcal{X},\mathcal{Y}).
\end{eqnarray}

\subsection{Deep cyclic mapping}

Following the functional map convention, given $C,\Phi,\Psi$ the \textit{soft correspondence matrix} mapping between $\mathcal{X}$ to $\mathcal{Y}$ reads \cite{fmnet}
\begin{eqnarray}
\label{eq:P_soft_map}
P = \bigl\lvert \Phi C \Psi^T \bigr\rvert
_{\mathcal{F}_{c}},
\end{eqnarray} 
where each entry $P_{ji}$ is the probability point $j$ in $\mathcal{X}$ corresponds to point $i$ in $\mathcal{Y}$.
We further use $\lvert\cdot\rvert_{\mathcal{F}_{c}}$ notation for column normalization, to emphasize the statistical interpretation of $P$.

Let $P$ represents the forward mapping $\pi_{\to}$ soft correspondence and $\tilde P$ the backward mapping $\pi_{\gets}$.
The cyclic distortion is defined by

\begin{equation} \label{eq:cyclic_loss}
\mathcal{L}_{cyclic}(\mathcal{X,Y}) = \\
\frac{1}{|\mathcal{X}|^2} 
\norm[\Bigg]{\biggl(D_\mathcal{X} - (\tilde{P} P)D_{\mathcal{X}}(\tilde{P} P)^{T}\biggr) }^2_{\mathcal{F}},
\end{equation} 
where $|\mathcal{X}|$ is the number of samples point pairs on $\mathcal{X}$.

Note that if we assumed the shapes were isometric, then we would have expected $D_\mathcal{Y}$ to be similar or even identical to $PD_\mathcal{X}P^T$, which yields once plugged into (\ref{eq:cyclic_loss}) the isometric constraint
\begin{equation} \label{eq:paper_loss}
\mathcal{L}_{isometric}(\mathcal{X,Y}) = \\
\frac{1}{|\mathcal{X}|^2} 
\norm[\Bigg]{\biggl(D_\mathcal{X} - \tilde{P} D_{Y}\tilde{P}^{T}\biggr) }^2_{\mathcal{F}}.
\end{equation} 
\noindent
The cyclic distortion (\ref{eq:cyclic_loss}) is self-supervised, as no labels are required, and only use the pair-wise distances on the source manifold $\mathcal{X}$. The conjugate mappings are based on the functional maps architecture and use the geometry of both spaces, source and target. Since we constraint the mapping on the source's geometry, the mapping copes with stretching, and thus learning invariant representations.

\section{Implementation}

\subsection{Hardware}
The network was developed in TensorFlow \cite{tensorflow2015-whitepaper}, running on a GeForce GTX 2080 Ti GPU. The SHOT descriptor~\cite{shotdesc} was implemented in MATLAB, while the Laplace Beltrami Operator (LBO)~\cite{LBO} and geodesic distances were calculated in Python.

\subsection{Pre-processing}
We apply a sub-sampling process for shapes with more than 10,000 points using qSlim~\cite{qSlim} algorithm.
SHOT descriptor was computed on the sub-sampled shapes, generating a descriptor of length $s=350$ per vertex. Finally, the LBO eigenfunctions corresponding to the least significant 70 eigenvalues were computed for each shape.
The distance matrices were computed using the Fast Marching algorithm~\cite{fastMarch}.
In order to initialize the conjugate mapping, we found that a hard constraint on $P$ and $\tilde P$ coupling provides good results. Specifically we minimized in the first epoch the cost function $||P\tilde{P} - I||^2_{\mathcal{F}}$ before applying the soft cyclic criterion (\ref{eq:cyclic_loss}).

\subsection{Network architecture}
The architecture is motivated by \cite{fmnet,halimi2019unsupervised} and shown in Figure \ref{fig:architecture_scheme}. 
The input to the first layer is the raw 3D triangular mesh representations of the two figures given by a list of vertices and faces. We apply a multi-layer SHOT~\cite{shotdesc} descriptor by evaluating the SHOT on $m \sim 5 $ global scaled versions of the input. The figures vary from 0.2 to 2 times the size of the original figures, followed by a $1$x$1$ convolution layers with $2m$ filters, to a $1\times1$ convolution layer with one filter, generating an output of $n\times s$ descriptor to the network.
Besides, the relevant eigenfunctions and pair-wise distance matrix of the source shape are provided as parameters to the network.

The next stage is the ResNet~\cite{Resnet} layers with the shared weights applied to both figures.
Subsequently, the non-linear descriptors are multiplied by the $n\times k$ LBO eigenfunctions.
We calculate the forward and backward mappings $C$ and $\tilde C$ using the same network and evaluate the corresponding forward and backward mappings $P$ and $\tilde P$, which are fed into the soft cyclic loss (\ref{eq:cyclic_loss}).

\section{Experiments}
In this section, we present multiple experiments in different settings; synthetic and real layouts, transfer learning tasks,  non-isometric transformations, partial matching and one-shot learning.
 We show benchmarks, as well as comparisons to state-of-the-art solutions, both for axiomatic and learned algorithms.
\subsection{Mesh error evaluation}
The measure of error for the correspondence mapping between two shapes will be according to the Princeton benchmark \cite{geodesic_error_metric}, that is, given a mapping $\pi_{\to}(\mathcal{X,Y})$ and the ground truth $\pi^*_{\to}(\mathcal{X,Y})$ the error of the correspondence matrix is the sum of geodesic distances between the mappings for each point in the source figure, divided by the area of the target figure.
\begin{equation} \label{eq:geodesic_error}
\epsilon(\pi_{\to}) = \sum_{x \in \mathcal{X}} \frac{\mathcal{D_Y}(\pi_{\to}(x),\pi^*_{\to}(x))}{\sqrt{area(\mathcal{Y})}},
\end{equation}
where the approximation of $area(\mathcal{\bullet})$ for a triangular mesh is the sum of it's triangles area.

\subsection{Synthetic FAUST}
We compared our alignment on FAUST dataset \cite{FAUST} versus supervised  \cite{fmnet} and unsupervised  \cite{halimi2019unsupervised} methods.
We followed the experiment as described in ~\cite{fmnet} and used the synthetic human shapes dataset, where the first 80 shapes (8 subjects with 10 different poses each) are devoted to training, and 20 shapes made of 2 different unseen subjects are used for testing.
For a fair comparison between methods, we did not run the PMF cleanup filter ~\cite{PMF} as this procedure is extremely slow and takes about 15 minutes for one shape build of $\sim 7k$ vertices on an i9 desktop.

We do not perform any triangular mesh preprocessing on the dataset, that is, we learn on the full resolution of 6890 vertices.
Each mini-batch is of size 4 (i.e 4 pairs of figures),  using $k = 120$ eigenfunctions, and  10 bins in SHOT with a radius set to be $5\%$ of the geodesic distance from each vertex.

We report superior results for inter-subject and intra-subject tasks in Table \ref{table:real_scal_results}, while converging faster (see Figure \ref{fig:different_losses}).

\subsection{Real scans}
We tested our method on real 3D scans of humans from the FAUST~\cite{FAUST} dataset. While the synthetic samples had $\sim 6k$ vertices, each figure in this set has $\sim 150k$ vertices, creating the amount of plausible cyclic mappings extremely high. The dataset consists of multiple subjects in a variety of poses, where none of the poses (e.g., women with her hands up) in the test set were present in the training set. The samples were acquired using a full-body 3D stereo capture system, resulting in missing vertices, and open-end meshes.
The dataset is split into two test cases as before, the intra and inter subjects (60 and 40 pairs respectively), and ground-truth correspondences in not given. Hence, the geodesic error evaluation is provided as an online service. 
As suggested in \cite{fmnet}, after evaluating the soft correspondence mappings, we input our map to the PMF algorithm \cite{PMF} for a smoother bijective correspondence refined map.
We report state of the art results on both inter and intra class mappings in comparison to all the unsupervised techniques.
We provide visualization in Figure \ref{fig:faust_real_scans} and qualitative results in table \ref{table:real_scal_results}.

\subsection{Non-isometric deformations}
An even bigger advantage of the proposed method is its ability to cope with local stretching. 
Due to the cyclic mapping approach, we learn local matching features directly between the models and are not relying on a base shape in the latent space or assume isometric consistency.
We experimented with models generated in Autodesk Maya that were locally stretched and bent.
We show visual results in Figure \ref{fig:local_scaling}. The proposed approach successfully handle large non-isometric deformations.

\subsection{TOSCA}\label{subsection:tosca}
The TOSCA dataset ~\cite{tosca_dataset} consists of 80 objects from different domains as animals and humans in different poses.
Although the animals are remarkably different in terms of LBO decomposition, as well as geometric characteristics, our model achieves excellent performance in terms of a geodesic error on the dataset after training for a single epoch on it, using the pre-trained model from the  real scans FAUST dataset.

In Figure \ref{fig:tosca_geo_error}, we show a comparison between our and other supervised and unsupervised approaches and visualize a few samples in Figure \ref{fig:tosca_visualization}. Compared methods results were taken from 
\cite{halimi2019unsupervised}.
Our network was trained for a single epoch on the dataset, with a pre-trained model of the real scans FAUST data and yet, shows great performance. We report state of the art results, compared to axiomatic, supervised, and unsupervised methods.   
Also note that while other methods mention training on each class separately, we achieve state-of-the-art results while training jointly.

\subsection{SCAPE} 

 To further emphasize the generalization capabilities of our network, we present our results on the SCAPE dataset ~\cite{scape_dataset}, which is an artificial human shape dataset, digitally generated, with completely different properties from the FAUST dataset in any aspect (geometric entities, scale, ratio, for example). Nevertheless, our network that was trained on the real scan FAUST dataset performs remarkably well. See Figure ~\ref{fig:scape_geo_error}.
  Compared methods results were taken from 
\cite{halimi2019unsupervised}.

\subsection{One-shot single pair learning}\label{subsection:one_shot_single_pair}
Following the experiment shown in  \cite{halimi2019unsupervised}, we demonstrate that
we can map in between two shapes seen for the first time without training on a large dataset.
Compared to optimization approaches we witness improved running time due to optimized hardware and software dedicated to deep learning in recent years. In Figure \ref{fig:single_pair_learning} we show such a mapping in between highly deformed shapes, and we found it intriguing that a learning method based on just two samples can converge to a feasible solution even without strong geometric assumptions. Note that in that case methods based on isometric criterion fail to converge due to the large non-isometric deformation.
In this experiment we used our multi-SHOT pre-trained weights before we ran our cyclic mapper.

\subsection{Partial shapes correspondence}
The partial shapes correspondence task is inherently more complicated than the full figure alignment problems. While in most experiments shown above, the number of vertices in both shapes differed by less than 5\%, in the partial shapes task, we consider mappings between objects that differ by a large margin of up to 75\% in their vertex count.
To this end, numerous bijective solutions, such as  \cite{sphericalbijective,kernelmaching,PMF} show degraded performance on the partial challenge, resulting in targeted algorithms    \cite{partial_maps,partial2} for the mission. With that in mind, we show our results on the SHREC 2016 ~\cite{cosmo2016shrec} partial shapes dataset. We use the same architecture as described earlier, given hyperparameters and trained weights from the TOSCA~\ref{subsection:tosca} experiment, showing our network's generalization capabilities. As before, we have trained the network on this dataset only for a single epoch.

\section{Limitations}
Our method uses functional maps architecture, which requires us to pre-compute sets of bases functions. To that end, this process can not be done in real-time in the current setup, and there might be an inconsistency in bases functions between shapes due to noise or large non-isometric deformations.
While this method works well for isometric or stretchable domains, once the deformations are significantly large, we found that the current system does not converge to a reasonable geodesic error in terms of a pleasant visual solution, which makes it challenging to use in cross-domain alignments.
We believe that the proposed approach can be used as part of semantic-correspondence to overcome those limitations.

\section{Summary}
We presented here a cyclic architecture for dense correspondence between shapes. 
This approach is self-supervised, can cope with local stretching as well as non-rigid isometric deformations.  It outperforms other unsupervised and supervised approaches on tested examples, and we report state-of-the-art results in several scenarios, including real 3D scans and partial matching task.


\clearpage
\nocite{Boscaini2016AnisotropicDD,masci2015geodesic,litman2013learning,SGMDS_axiomatic,rodola2014dense}

{\small
\bibliographystyle{ieee_fullname}
\bibliography{egbib}
}

\end{document}